\PassOptionsToPackage{table}{xcolor}
\documentclass[11pt,a4paper]{article}
\usepackage[hyperref]{dert}
\usepackage{amsmath}
\usepackage{tikz}
\usepackage{pgfplots}
\usepackage{times}
\usepackage{latexsym}
\usepackage{multirow}
\usepackage{url}
\usepackage{verbatim}
\usepackage{xcolor}

\aclfinalcopy % Uncomment this line for the final submission
\def\Baseline{Transformer}
\def\GenericInit{\langle params_{enc}, params_{dec}\rangle}
\def\RandInit{\langle \mathrm{Random}, \mathrm{Random}\rangle}
\def\BertRandInit{\langle \mathrm{Bert}, \mathrm{Random}\rangle}
\def\BertBertInit{\langle \mathrm{Bert}, \mathrm{Bert}\rangle}

\def\CnndmCnndmInit{\langle \mathrm{CnnDm_{enc}}, \mathrm{CnnDm_{dec}}\rangle}
\def\GwordGwordInit{\langle \mathrm{Gword_{enc}}, \mathrm{Gword_{dec}}\rangle}
\def\BCBCInit{\langle \mathrm{Bert\&CnnDm_{enc}}, \mathrm{Bert\&CnnDm_{dec}}\rangle}
\def\BGBGInit{\langle \mathrm{Bert\&Gword_{enc}}, \mathrm{Bert\&Gword_{dec}}\rangle}

\title{Multi-stage Pretraining for Abstractive Summarization}

\author{Sebastian Goodman \\
  Google Research \\
  \texttt{seabass@google.com} \\\And
  Zhenzhong Lan \\
  Google Research \\
  \texttt{lanzhzh@google.com} \\\And
  Radu Soricut \\
  Google Research \\
  \texttt{rsoricut@google.com} \\}

\date{}

\begin{document}
\maketitle

% Abstract
\begin{abstract}
Neural models for abstractive summarization tend to achieve the best performance in the presence of highly specialized, summarization specific modeling add-ons such as pointer-generator, coverage-modeling, and inference-time heuristics.
We show here that pretraining can complement such modeling advancements to yield improved results in both short-form and long-form abstractive summarization using two key concepts: full-network initialization and multi-stage pretraining.
Our method allows the model to transitively benefit from multiple pretraining tasks, from generic language tasks to a specialized summarization task to an even more specialized one such as bullet-based summarization.
Using this approach, we demonstrate improvements of 1.05 ROUGE-L points on the Gigaword benchmark and 1.78 ROUGE-L points on the CNN/DailyMail benchmark, compared to a randomly-initialized baseline.
\end{abstract}

% Introduction
\section{Introduction}
The field of abstractive summarization has exploded since the introduction of neural models for text generation, as such models have been successful at pushing the state-of-the-art ever higher~\cite{rush2015neural,luong2015multi,nallapati2016abstractive,see2017get,paulus2017deep,amplayo2018entity,gehrmann2018bottom}.
One of the main attractions for some of these neural models is that they are capable of generating summaries from scratch, based on abstract representations of the input document, and therefore promise to allow for the abstraction that is intuitively needed when compressing longer pieces of text into shorter ones.
However, there are two main trends that we have observed in multiple modeling approaches.

First, models that are successful at improving the state-of-the-art are usually ones that take a generic deep neural-network architecture as a base model and add various specialized, summarization-specific learning mechanisms and inference heuristics.
Among such mechanisms are the ability to copy from the input~\cite{gu2016incorporating,see2017get} (i.e., switch between extractive and abstractive styles of summarization) and the ability to model coverage~\cite{suzuki2016rnn,see2017get,gehrmann2018bottom} (mechanisms that specifically target the ability to learn what to cover and what to ignore from the input).
Inference-time heuristics include length penalties, coverage penalties, and n-gram--based repeat restrictions enforced by the decoder~\cite{gehrmann2018bottom}.
Recent works such as~\cite{holtzman2019curious,welleck2019neural} propose alternative training objectives which lead to better quality neural text generation systems by reducing repetitions and blandness.

Second, the push towards improved quantitative results, as measured by various ROUGE-based metrics~\cite{lin2004rouge}, tends to diminish the extent to which these models generate summaries that are truly abstractive.
Instead, the models learn to generate fluent outputs based on extracting/copying the right words and phrases from the input, in effect a small-granularity extractive process.
To quantify this better, we mention here that for the CNN-DailyMail dataset~\cite{hermann2015teaching}, the abstraction rate (percent of words in the summary that are not present in the input) is around 14\%, while a current state-of-the-art model produces outputs with an abstraction rate of 0.5\%~\cite{gehrmann2018bottom}.

We present in this paper an approach to abstractive summarization that diverges from the two trends mentioned above, by addressing the problem in a data-driven way.
Through pretraining, we improve the model's fundamental language capabilities, in a way that is complementary to the modeling advancements described above.

Pretraining has become a topic of great interest since the introduction of the BERT~\cite{devlin2018bert} model for language understanding and GPT-2~\cite{radford2019language} for neural text generation.
Works such as~\cite{edunov2019pre,zhang2019pretraining} show that pre-trained networks can be used to improve performance on summarization benchmarks.
In~\cite{rothe2019leveraging}, they demonstrate that pre-trained networks can be used to improve performance on a variety of tasks, beyond just summarization.
In this work, we explore different initialization schemes for a Transformer~\cite{vaswani2017attention} model and show how they affect performance on summarization tasks, after fine-tuning.
Compared to a randomly-initialized baseline, this improves upon the widely-used Gigaword~\cite{ldc-english-gigaword} benchmark by 1.05 ROUGE-L points.

Furthermore, we introduce multi-stage pretraining as a novel way to leverage multiple pretraining sources when solving a particularly hard problem, such as the non-anonymized CNN/DailyMail task~\cite{hermann2015teaching,nallapati2016abstractive}.
By taking the parameters of the state-of-the-art Gigaword model described above, and using them to initialize a model for the CNN/DailyMail task, we improve upon a randomly-initialized baseline by 1.78 ROUGE-L points.
The resulting model combines summarization-specific modeling advances with the benefits of multi-stage pretraining to achieve a model with a substantially higher abstraction rate than a comparable model without pretraining: in contrast with the 0.5\% rate of~\citet{gehrmann2018bottom}, our model outputs have an abstraction rate of around 4\%.
This difference indicates that there are substantive differences between the levels of abstractization achieved.

% Related work
\section{Related Work}
Early summarization work~\cite{dorr2003hedge,jing1999decomposition, filippova2015sentence} mostly focused on purely extractive approaches.
With the development of neural sequence-to-sequence models~\cite{sutskever2014sequence,bahdanau2015neural}, there has been substantial work in abstractive approaches~\cite{rush2015neural,luong2015multi,nallapati2016abstractive,chopra-etal:2016,see2017get,paulus2017deep,chen2018fast}.
However, due to the large search space and the relative lack of training data, purely abstractive approaches still suffer from performance issues compared to extractive approachs, especially from issues like missing critical details and hallucinations~\cite{hsu2018unified}.
A natural step is to combine the abstractive and extractive apporaches together.
Copy mechanisms~\cite{vinyals2015pointer, gu2016incorporating, see2017get} and coverage modeling~\cite{suzuki2016rnn,chen2016distraction,see2017get,gehrmann2018bottom} are examples of these approaches.

Our work represents a more fundamental way of solving the data-hungry problem and under-constrained modeling problem.
Instead of constraining the learning space, we incrementally pretrain the model on relatively similar, data-rich tasks.
Pretraining has been popular for image classification~\cite{he2018rethinking}, language understanding, see most recently~\cite{devlin2018bert}, and language generation~\cite{ramachandran2017unsupervised,edunov2019pre,rothe2019leveraging} tasks.
Although recent works in image classification~\cite{he2018rethinking} show that pretraining may not be necessary when are allowed to train for a longer time,
our results indicate that meaningful multi-stage pretraining allows our models to achieve significantly higher accuracy levels.

% Datasets
\section{Datasets}

\begin{table*}[t]
\small
\centering
\begin{tabular}{ l | c c c | c c | c } %% | c c c c }
                  & Training  & Dev      & Test     & Input     & Output  &  Abstraction \\
                  & Examples  & Examples & Examples & Length    & Length  &  Rate\% \\
  \hline
  Gigaword        & 4,194,451         &  10,000        &  1,951         &          37.4 &               9.8 &       53.8\% \\
  CNN/DailyMail   &   287,108         &  13,368        & 11,489         &         572.7 &              66.0 &       13.8\%
\end{tabular}
\caption{Statistics for the Gigaword and CNN/DailyMail datasets.
  %The first three columns show the number of examples in the train, dev, and test sets. % Can-be-deleted
  The fourth and fifth column show the input/output sequence length over the dev set (number of word-pieces).
  The final column shows Abstraction Rate, the percentage of tokens in the reference summary not present in the input.
  %For the Gigaword test set, we use the official one provided by \cite{rush2015neural}. % Can-be-deleted
}
\label{table:datasets}
\end{table*}

\begin{table}
\small
\centering
\begin{tabular}{ l | c c | c c } %% | c c c c }
                  & Input & Output & Input   & Output  \\
                  & Limit & Limit  & Trunc.\% & Trunc.\% \\
  \hline
  Gigaword        &            128 &              64 &    .07\% &    0\% \\
  CNN/DailyMail   &            640 &              96 &  64.95\% & 19.6\% \\
\end{tabular}
\caption{The left two columns show the sequence length limits we use (number of word-pieces).
  The right two columns show the percentage of examples in the dev set that were truncated as a result of these limits (in the Gigaword case, the data are pre-truncated; the truncation rates given here are in addition to that).}
\label{table:truncation}
\end{table}

We use two News summarization datasets in our experiments: the Gigaword~\cite{rush2015neural} and the CNN/DailyMail~\cite{nallapati2016abstractive} datasets.
The Gigaword dataset contains examples of $\langle$document, headline$\rangle$ as input/output pairs, while the CNN/DailyMail dataset contains examples of $\langle$document, bullet-summary$\rangle$ as input/output pairs.
Gigaword represents a short-form summarization task, in which the output is confined to a single sentence (i.e., headline).
CNN/DailyMail, on the other hand, represents a long-form summarization task, in which the output is relatively long and contains multiple sentences (i.e., bullet-based summary sentences).

Given that these datasets share the same domain (i.e., News reported in English), it is reasonable to expect that general concepts learned on one dataset transfer to the other.
Nevertheless, there are also significant differences between them: CNN/DailyMail has far fewer training examples than Gigaword (about 20x fewer), and much longer inputs (15x longer on average) and outputs (7x longer on average), see Table~\ref{table:datasets}.
However, it is intuitively appealing to train a long-form summarization model (on CNN/DM) by first pretraining it on for short-form summarization (on Gigaword).
Moreover, one should not run before learning to walk; i.e., short-form summarization should not start from scratch, but from a model that is already equipped for natural language understanding.,
To that end, we start from a BERT model pretrained on the BooksCorpus~\cite{zhu2015aligning} and English Wikipedia, as described in~\cite{devlin2018bert}.
As we see in Sec.~\ref{sec:experiments}, we achieve significant improvements using this style of multi-stage pretraining.

\section{Models and Initialization Schemes}
In this section, we describe key aspects of our summarization models and the initialization strategies we use.

\subsection{Base Model}
Our base model is a Transformer Network~\cite{vaswani2017attention} in a BERT configuration~\cite{devlin2018bert}.
We chose the BERT model configuration for studying multi-stage pretraining because
i) BERT models trained on unsupervised language tasks are readily available, and
ii) models based on BERT parameters have been shown to perform very well on a large number of language understanding tasks.

The Transformer model consists of two components, an Encoder and a Decoder.
The role of the Encoder is to map input tokens to embeddings to token-in-context representations, in this case using the self-attention mechanism.
The role of the Decoder is to predict a current output token given the previous outputs and the encoded input representations, using both self-attention and encoder-decoder attention mechanisms.

Purely abstractive approaches model predictions as a distribution across the token vocabulary at each timestep $t$:
\begin{align}
P(t) = P_{vocab}(t) &= softmax(\hat{y}_t) \label{eq:predict} \\
                    &= softmax(Vd_t + b_v) \nonumber
\end{align}
\noindent
where $V$ is a learnable embedding matrix.
Note that we use the same embedding matrix for also embedding the input tokens.
We use MLE as our training objective (negative log likelihood of the target token).
We extend Eq.\ref{eq:predict} to models with copy-attention below.

\begin{comment} % There is nothing new here, we can skip it.
We use the negative log likelihood of the target token $w^{*}_t$ as our training objective:
\begin{align}
loss_t &= -log(P_{w^{*}_t}(t))
\end{align}

To compute the batch loss, we average across both batch and sequence dimensions (non-pad positions only):
\begin{align}
loss &= \frac{1}{T}\sum_{t=0}^{T}loss_t
\end{align}
\end{comment}

\subsection{Copy-attention}
Mechanisms for copying words from the input such as the pointer-generator network~\cite{see2017get} have helped push forward the state-of-the-art in abstractive summarization.
We explore an approach here similar to the CopyTransformer~\cite{gehrmann2018bottom}, where a single attention head in the encoder-decoder attention is used as the copy distribution.

Our approach diverges slightly from previous approaches by modeling copying and generation as a single event.
The pointer-generator network~\cite{see2017get}, for example, augments the purely abstractive predictions of Eq.\ref{eq:predict} with an extractive component using a learned weighted sum of generation and copy probabilities:
\begin{align*}
P(t) = p_{gen}(t)P_{vocab}(t) + (1-p_{gen}(t))a_tX
\end{align*}
\noindent
In this formulation, copy probabilities are computed by projecting the encoder-decoder attention probabilities $a_t$ into the vocabulary space using a one-hot encoding of the input ids, $X$.

Our approach is similar to the formulation above but uses a learned weighted sum of logits, normalizing afterwards:
\begin{align}
\hat{z}_t &= p_{gen}(t)\hat{y}_t + (1-p_{gen}(t))\hat{a}_tX
\label{eq:pgen}
\end{align}
\begin{align*}
P(t) &= softmax(\hat{z}_t)
\end{align*}
\noindent
This change preserves the model's ability to copy but does not use distinct probabilities for copying and generation.
Furthermore, it allows the objective function to be simplified by distributing the logarithm into the softmax.

In our approach, the learned weight function, $p_{gen}(t)$, is implemented as a fully connected layer on top of decoder output $d_t$:
\begin{align}
p_{gen}(t) &= \sigma(x_g^Td_t + b_g)
\end{align}
where $x_g$ and $b_g$ are learnable parameters of the model.

\subsection{Content selection}
Bottom-up attention uses a content selection model to restrict the copy attention to input tokens predicted to appear in the output.
This technique has been shown to improve summarization performance~\cite{gehrmann2018bottom}.

We take a similar approach but instead use a content selection model that is based on BERT fine-tuning~\cite{devlin2018bert}.
To train the content selection model, we construct labels by aligning the input document and groundtruth summary, according to the procedure described in ~\cite{gehrmann2018bottom}.

We model the probability of selecting token $i$, $P_{sel}(i)$, using a fully connected layer on top of the BERT model's outputs $c_i$:
\begin{align*}
P_{sel}(i) &= \sigma(x_s^Tc_i + b_s)
\end{align*}

We use logistic regression to train the content selection model (on non-pad positions only).
Finally, we integrate the trained content selection model into the summarization model by masking the encoder-decoder attention logits (copy head only) in Eq.\ref{eq:pgen}:
\begin{align}
  \hat{a}'_t =
\begin{cases}
    \hat{a}_t,& \text{if }P_{sel}(i) > \epsilon \\
    \hat{a}_t - 10000,& \text{otherwise}
\end{cases}
\end{align}
\noindent
The threshold $\epsilon$ is chosen by sorting the model's predictions over the dev set, and choosing the midpoint that maximizes F1 score.

In our experiments, we also use an oracle version, which uses the constructed labels $y_i \in \{0,1\}$ instead of the predicted probabilities $P_{sel}(i)$.

\subsection{Initialization Schemes}
We present here the initialization schemes we use in this paper, namely zero-step, one-step, and two-step initializations.
We use the notation $\GenericInit$ to represent the parameter initialization for the encoder and decoder, respectively.

\textbf{Zero-step} initialization means that the model parameters are not pretrained, i.e., they are initialized from random distributions.
In our experiments, we use the default distributions provided with the BERT model.
Under our notation, we denote this initialization as $\RandInit$.

\textbf{One-step} means that some of the parameters of the model have been pretrained.
We use the following one-step initialization schemes:

\begin{description}
\item[$\BertRandInit$]: encoder parameters are initialized from the BERT checkpoint; decoder parameters are initialized randomly.
\item[$\BertBertInit$]: encoder and decoder initialized symmetrically from the BERT checkpoint (cross-attention initialized from self-attention).
\item[$\GwordGwordInit$]: encoder and decoder initialized from parameters resulted from training end-to-end on the Gigaword corpus (initialized using $\RandInit$).
\item[$\CnndmCnndmInit$]: encoder and decoder initialized from parameters resulted from training end-to-end on the CNN/DM corpus (initialized using $\RandInit$).
\end{description}

\textbf{Two-step} means that the model parameters have been pretrained in a two-step fashion:
first on one task, followed by another task (as a fine-tune procedure).

\begin{description}
\item[$\BCBCInit$]: encoder and decoder initialized from the result of: i) starting from $\BertBertInit$ initialization, and ii) fine-tuning on the CNN/DM corpus.
\item[$\BGBGInit$]: encoder and decoder initialized from the result of: i) starting from $\BertBertInit$ initialization, and ii) fine-tuning on the Gigaword corpus.
\end{description}
\noindent
In turn, a two-step initialization such as $\BGBGInit$ can be used as a starting point for some subsequent fine-tuning training procedure, for instance on the CNN/DM corpus.
It is important to note that the initialization on the encoder side is different from the one on the decoder side, as the parameters of the two networks specialize in their different roles:
one for encoding representations and the other one for decoding representations.
The experimental results in the following section indicate that chaining fine-tuning procedures in this manner makes learning more effective.

% Experimental Results
\section{Experiments}
\label{sec:experiments}

In this section, we describe our experimental setup (hardware, model hyperparameters) and our experimental results.

\subsection{Implementation details}
All models are trained on Google Cloud TPUs with 16GB high-bandwidth memory (HBM) each.
The models are evaluated on Tesla GPUs, as the TPUs do not support some of the string operations used for evaluation.
We use TensorFlow~\cite{tensorflow2015-whitepaper} and a patched version of the BERT model~\cite{devlin2018bert}.

All BERT checkpoints used are provided by~\citet{devlin2018bert}.
We limit ourselves to the \textit{uncased\_L-12\_H-768\_A-12} monolingual version.
Larger versions are available, but they become too memory intensive when used in a dual Transformer encoder-decoder setup like ours.

For the experiments presented here, we preprocess both datasets by tokenizing into word-pieces, as described in~\cite{devlin2018bert}.
Due to hardware and architecture limitations, we truncate (or pad) inputs and outputs to a fixed number of word-pieces, as shown in Table~\ref{table:truncation} (these limits are also reflected in Table~\ref{table:datasets}).

\subsection{Hyperparameters}
All of our models use the Adam optimizer and learning rate of $2e^{-5}$.
%which is close to the learning rate that is recommended for fine-tuning by\cite{devlin2018bert} (i.e. $1e^{-5}$).
We experimented with three learning rate values: $1e^{-5}$, $2e^{-5}$, and $5e^{-5}$.
We found that $1e^{-5}$ does not improve the results compared to $2e^{-5}$, while $5e^{-5}$ resulted in poor performance for the $\RandInit$ initialization scheme.

We use .3 dropout on all Transformer layer outputs.
Dropout is done at a token level, in keeping with the recommendations from~\cite{devlin2018bert} and~\cite{vaswani2017attention}.
We tested four values for the dropout rate: .1, .2, .3, and .4.
We found that a dropout rate of .3 worked best for both tasks.

Our vocabulary size is 30,522 word pieces, matching the vocabulary provided with the BERT checkpoint.
In addition, the version of BERT we use has 12-layers, each with a size of 768.

For experiments on the Gigaword dataset, our model has 128 encoder positions and 64 decoder positions (see Table~\ref{table:truncation}).
For experiments on the CNN/DM dataset, our model has 640 encoder positions and 96 decoder positions.

On Gigaword, we use beam search decoding with beam width 4 and length penalty parameter $\alpha=.6$~\cite{wu2016gnmt}.
Greedy decoding performed slightly worse across all setups by about $.2$ ROUGE-L F1 points.
On CNN/DM, we use greedy decoding only, as we found beam search yields similar performance.

\begin{table}
\small
\centering
\begin{tabular}{ l | c c } %% | c c c c }
                  % & \multicolumn{2}{ |c } { AUC } %% & \multicolumn{4}{ |c }{ Classification }
                  & AUC-PR & AUC-RoC               \\ %% & $\epsilon$ & Precision & Recall & F1
  % \hline
  % Gigaword dev  &  78.33 &   91.42                 \\ %% & \multirow{2}{*}{.376} & 66.76 & 71.46 & 69.03
  % Gigaword test &  68.91 &   87.70                 \\ %% &                       & 56.06 & 64.78 & 61.31
  \hline
  CNN/DM dev    &  82.43 &   87.86                 \\ %% & \multirow{2}{*}{.425} & 70.08 & 82.36 & 75.72
  CNN/DM test   &  81.90 &   87.84                 \\ %% &                       & 68.98 & 82.73 & 75.23
\end{tabular}
\caption{AUC of our BERT-based content selection model.
Note that these figures are not directly comparable to \cite{gehrmann2018bottom} due to differences in tokenization.}
\label{table:cauc}
\end{table}

\begin{table}
\small
\centering
\begin{tabular}{ l | l | c c c }
                                &          & Precision &  Recall & F1 \\
  % \hline
  % \multirow{2}{*}{Gigaword dev}  & Oracle  & 100.00    &   70.61 &  81.67 \\
  %                                & Model   &  68.42    &   44.97 &  52.89 \\
  % \hline
  % \multirow{2}{*}{Gigaword test} & Oracle  & 100.00    &   58.12 &  71.02 \\
  %                                & Model   &  58.51    &   33.81 &  40.68 \\
  \hline
  \multirow{2}{*}{CNN/DM dev}    & Oracle  & 100.00    &   80.77 &  89.01 \\
                                 & Model   &  56.11    &   46.47 &  49.95 \\
  \hline
  \multirow{2}{*}{CNN/DM test}   & Oracle  & 100.00    &   80.49 &  88.85 \\
                                 & Model   &  55.11    &   46.58 &  49.58 \\
\end{tabular}
\caption{Label coverage of our BERT-based content selection model. We also show
the performance of a content selection oracle which always does perfect
content selection. True positives here are groundtruth word pieces that are selected
by the content selector. The oracle achieves perfect precision because the labels
are used to select inputs in the first place. The oracle does not achieve perfect recall
because not all input word pieces are present in the groundtruth.}
\label{table:coracle}
\end{table}

\begin{table*}[t]
\rowcolors{2}{gray!10}{white}
\small
\centering
\begin{tabular}{ r | c c c | c }
                                                  Method & \multicolumn{3}{ |c| } { ROUGE F1 }               & Best (on dev) \\
                                                         & 1               & 2              & L              & at epoch      \\
   \hline
   ABS+~\cite{rush2015neural}                            & 29.78           & 11.89          & 26.97          & - \\
   Luong-NMT~\cite{luong2015multi}                       & 33.10           & 14.45          & 30.71          & - \\
   Feat2s~\cite{nallapati2016abstractive}                & 32.67           & 15.59          & 30.64          & - \\
   RAS-Elman~\cite{chopra2016abstractive}                & 33.78           & 15.97          & 31.15          & - \\
   SEASS~\cite{zhou2017selective}                        & 36.15           & 17.54          & 33.63          & - \\
   Re\textsuperscript{3}Sum~\cite{cao2018retrieve}                        & 37.04           & 19.03          & 34.46          & - \\
   Base+E2Tcnn+sd~\cite{amplayo2018entity}               & 37.04           & 16.66          & 34.93          & - \\
   \hline
   \Baseline $\RandInit$                                  & 38.05          & 18.95          & 35.26          & 68 \\
   \Baseline $\BertRandInit$                              & 38.84          & 19.86          & 36.24          & 47 \\
   \Baseline $\BertBertInit$                              & 38.96          & 19.55          & 36.22          & 88 \\
  % \Baseline $\CnndmRandInit$                             & 38.38          & 19.38          & 35.66          & 72 \\
   \Baseline $\CnndmCnndmInit$                            & 38.79          & 19.88          & 36.14          & 47 \\
   \Baseline $\BCBCInit$                                  & \textbf{39.14} & \textbf{19.92} & \textbf{36.57} & 41  \\
   \hline
   +CopyTransformer $\RandInit$                        & 37.98          & 18.93          & 35.23          & 47 \\
   +CopyTransformer $\BertRandInit$                    & 38.94          & 19.91          & 36.24          & 34 \\
   +CopyTransformer $\BertBertInit$                    & 38.97          & 19.84          & 36.31          & 59 \\
  % +CopyTransformer $\CnndmRandInit$                   & 38.10          & 18.80          & 35.19          & 43 \\
   +CopyTransformer $\CnndmCnndmInit$                  & 38.55          & 19.44          & 35.88          & 51  \\
   +CopyTransformer $\BCBCInit$                        & 38.52          & 19.28          & 35.91          & 38  \\
   \hline
   +Bottom-Up CopyTransformer $\BertBertInit$                        & 38.90          & 19.70          & 36.28          & 34  \\
   +Bottom-Up CopyTransformer Oracle $\BertBertInit$               & 53.98*         & 32.97*          & 49.37*         & 91  \\
\end{tabular}
\caption{Results of summarization methods on the Gigaword benchmark.
  The first section shows the performance reported by prior work.
  The second section shows our baseline model performance under various initialization schemes.
  The third section shows the performance of our model with the addition of the copy-attention mechanism, while the final section shows the performance when using the content selection model in a two-step setup.}
  Scores marked with an asterisk represent results when using an oracle for content selection.
\label{table:gigaword}
\end{table*}

\subsection{Experimental Results}
To set up the end-to-end experimental conditions, we first present the results obtained by our content selection model.
We follow with end-to-end results on the Gigaword and CNN/DM benchmarks.
Finally, we present an ablation study on partial initialization.

\subsubsection{Content selection}
Table~\ref{table:cauc} shows that our BERT-based content selection model achieves an 80+ AUC score on the validation set, with similar performance on the test set.
Though our model's AUC seems on par with that of~\cite{gehrmann2018bottom}, we note that the comparison is imperfect due to differences in tokenization.

Table~\ref{table:coracle} quantifies the performance from another perspective, i.e. label coverage, using Recall/Precision/F1 metrics.
Label coverage is an important metric for understanding the performance of content selectors in the context of copy mechanisms.
If the content selector has a false negative on a label word, then that word cannot be copied, thus hurting performance.
Similarly, if the content selector has a false positive on a label word, the usefulness of the content selector model degrades.
Our content selection model seems sub-par compared to the oracle, with slightly less than half of the labels present in the content selector's outputs, and slightly more than half of the content selector's outputs present in the groundtruth labels.

Based on the results above, our content selection model should improve summarization performance when used in a two-step setup, as~\cite{gehrmann2018bottom} demonstrate with their Bottom-Up Summarization (CopyTransformer) method.
Our results on summarization benchmarks below, however, show that the actual improvement to performance on top of pretraining is minimal.

\subsubsection{The Gigaword Benchmark}

\begin{table*}[t]
\rowcolors{2}{gray!10}{white}
\small
\centering
\begin{tabular}{ r | c c c | c }
                                                Method & \multicolumn{3}{ |c| } { ROUGE F1 }              & Best (on dev) \\
                                                       & 1              & 2              & L              & at epoch \\
 \hline
 Pointer-Generator~\cite{see2017get}                   & 36.44          & 15.66          & 33.42          & - \\
 Pointer-Generator + Coverage~\cite{see2017get}        & 39.53          & 17.28          & 36.38          & - \\
 ML + Intra-Attention~\cite{paulus2017deep}            & 38.30          & 14.81          & 35.49          & - \\
 Bottom-Up Summarization~\cite{gehrmann2018bottom}     & 41.22          & 18.68 & 38.34          & - \\
 \hline
 ML + RL~\cite{paulus2017deep}                         & 39.87          & 15.82          & 36.90          & - \\
 Saliency + Entailment reward~\cite{pasunuru2018multi} & 40.43          & 18.00          & 37.10          & - \\
 Key information guide network~\cite{li2018actor}      & 38.95          & 17.12          & 35.68          & - \\
 Inconsistency loss~\cite{hsu2018unified}              & 40.68          & 17.97          & 37.13          & - \\
 Sentence Rewriting~\cite{chen2018fast}                & 40.88          & 17.80          & 38.54          & - \\
 SRC-ELMO+SHDEMB                                       & \textbf{41.56} & \textbf{18.94} & 38.47          & - \\
 \hline
 \Baseline $\RandInit$                                  & 39.20          & 16.39          & 36.49          & 177 \\
 \Baseline $\BertRandInit$                              & 40.00          & 17.11          & 37.35          & 143 \\
 \Baseline $\BertBertInit$                              & 40.67          & 17.50          & 37.90          & 232 \\
% \Baseline $\GwordRandInit$                             & 38.96          & 16.47          & 36.43          & 115 \\
 \Baseline $\GwordGwordInit$                            & 40.05          & 17.11          & 37.36          &  89 \\
 \Baseline $\BGBGInit$                                  & 40.80          & 17.90          & 38.14          &  72 \\
 \hline
 +CopyTransformer $\RandInit$                        & 39.43          & 16.58          & 36.82          & 135 \\
 +CopyTransformer $\BertRandInit$                    & 39.69          & 16.99          & 37.11          & 146 \\
 +CopyTransformer $\BertBertInit$                    & 40.78          & 17.73          & 38.18          & 186 \\
% +CopyTransformer $\GwordRandInit$                   & 40.24          & 17.53          & 37.69          & 111 \\
 +CopyTransformer $\GwordGwordInit$                  & 40.84          & 17.74          & 38.11          &  84 \\
 +CopyTransformer $\BGBGInit$                        & 41.20          & 18.13          & 38.57          &  83 \\
 \hline
 +Bottom-Up CopyTransformer $\BertBertInit$                        & 41.26 & 18.05          & \textbf{38.60} &  98 \\
 +Bottom-Up CopyTransformer Oracle $\BertBertInit$               & 62.59*         & 32.61*         & 57.34*         &  51 \\
\end{tabular}
\caption{Results of summarization methods on the CNN/DM benchmark.
  The first section shows the performance of models trained with MLE loss (directly comparable to ours).
  The second section shows the performance of Reinforcement-Learning--based approaches.
  The third section shows the performance of our base model under various initialization schemes.
  The fourth section shows the performance of our model with the addition of the copy-attention mechanism, while the final section shows the performance when using the content selection model in a two-step configuration.
  Scores marked with an asterisk represent results when using an oracle for content selection.
}
\label{table:cnndm}
\end{table*}

The results on the Gigaword benchmark summarization task are presented in Table~\ref{table:gigaword}.
Our best model reaches 36.57 ROUGE-L F1 on the test set, without using content selection or summarization-specific coverage penalties.
Overall, the results in Table~\ref{table:gigaword} indicate that the key components leading to improved performance on this task are:
(a) the large Transformer model,
(b) the full-network initialization of the model parameters and,
(c) the multi-stage pretraining scheme,
as discussed in detail below.

The ROUGE-L F1 score of 35.26 for the \Baseline $\RandInit$ model indicates that the learning capacity of the base \Baseline model is higher compared to the previous models.
On top of it, full-network (deep) initialization with the original BERT weights yields a +1 ROUGE-L increase, at 36.24 (for encoder-only) and 36.22 (encoder and decoder).
In contrast, deep initialization with weights originating from CNN/DM pretraining yield slightly lower results (36.14 ROUGE-L F1).
The best result is obtained with the $\BCBCInit$ initialization scheme (36.57 ROUGE-L F1), indicating that this multi-stage pretraining scheme helps the most with learning for this task, by first pretraining on generic language understanding tasks (the BERT stage), second pretraining on a related summarization task (CNN/DM), and finally fine-tuning on the target task (Gigaword).
We also report that, for these models, the abstraction rate (percent of words in the summary that are not present in the input) is in the range of 27-29\%.
While this is substantially less compared to the reference (at 53.8\%, see Table~\ref{table:datasets}), it still demonstrates non-trivial levels of abstractiveness.

We also note here that neither the copy-attention mechanism, nor its augmentation with content-selection prediction, improve on the results of the base model.
One possible explanation is that copy-attention is not well suited for a highly-abstractive task such as the one for the Gigaword benchmark.
This is supported by the result for +CopyTransformer $\BCBCInit$ model, which, at 35.91 ROUGE-L F1, indicates a significant quality degradation, possibly because of the pre-training on a highly-extractive task such as CNN/DM.

\subsubsection{The CNN/DailyMail Benchmark}

The results on the CNN/DM benchmark summarization task are presented in Table~\ref{table:cnndm}.

The first notable result is that the base model \Baseline $\RandInit$ achieves, at 36.49 ROUGE-L F1, a score that is on-par with the Pointer-Generator + Coverage of~\citet{see2017get} (but without any summarization-specific modeling).
Similar to the result on the Gigaword benchmark, deep (full-network) initialization with the original BERT weights for the encoder yields a +1 ROUGE-L increase, at 37.35 ROUGE-L F1, while deep initialization with BERT weights on both the encoder and decoder achieves an additional +0.5 increase, at 37.90 ROUGE-L F1.
Again, the multi-stage pretraining scheme \Baseline $\BGBGInit$, in which we first pretrain on generic language understanding (the BERT stage), then pretrain on a related summarization task (Gigaword), and finally fine-tuning on the target task (CNN/DM), yields the best result for base model, at 38.14 ROUGE-L F1.

We observe the same trends for the copy-attention model, but with significant performance increases relative to the baseline.
This is remarkable especially as this model is trained only with MLE loss, and does not suffer from the train-time inefficiencies present in RL-based approaches.
Also notably, the addition of the content-selection prediction yields almost no increase (from 38.57 to 38.60), in stark contrast with the large impact achieved by~\citet{gehrmann2018bottom} for their setup.
% It may be important for the content selection model and summarization encoder to use a different model, training data, or initialization.

In addition, our models score in the range of 4-5\% on the Abstraction-Rate metric.
While less than the reference (at 13.8\%, Table~\ref{table:datasets}), it is still much more novel than the best MLE model to date~\cite{gehrmann2018bottom}, measured at .5\% on this metric.
This suggests that our model is capable of better learning to maintain the balance between extractiveness and abstractiveness present in the data.
We emphasize here that our results do not make use of any of the inference heuristics (length penalty, coverage penalty, n-gram--based repeat restrictions) previously reported~\cite{gehrmann2018bottom} to be crucial for achieving high performance levels.

\begin{figure}
\centering
\resizebox{.9\width}{!}{%
  \begin{tikzpicture}
    \begin{axis}[
      xlabel=Number of pretrained layers,
      ylabel=ROUGE-L F1]
      \addplot[color=red,mark=x, error bars/.cd, y dir=both, y explicit,][domain=0:24] coordinates {
        (1,0.36245)  += (0, .00212) -= (0, .00202)
        (3,0.35483)  += (0, .00211) -= (0, .00202)
        (5,0.36064)  += (0, .00210) -= (0, .00199)
        (7,0.36857)  += (0, .00205) -= (0, .00203)
        (9,0.36776)  += (0, .00214) -= (0, .00208)
        (11,0.372)   += (0, .00207) -= (0, .00213)
        (13,0.36919) += (0, .00214) -= (0, .00211)
        (15,0.37873) += (0, .00204) -= (0, .00200)
        (17,0.38387) += (0, .00222) -= (0, .00209)
        (19,0.38152) += (0, .00212) -= (0, .00224)
        (21,0.38264) += (0, .00217) -= (0, .00212)
        (23,0.38329) += (0, .00209) -= (0, .00226)
      };
      \addplot[color=blue,dashed][domain=0:24]{.3716};
      \node at (axis cs:0,.3716) [anchor=south west] {$\RandInit$};
      \addplot[color=blue,dashed][domain=0:24]{.3849};
      \node at (axis cs:0,.3849) [anchor=south west] {$\BGBGInit$};
   \end{axis}
  \end{tikzpicture}%
}
\caption{Performance of the +CopyTransformer $\BGBGInit$ model on the CNN/DailyMail task with increasing numbers of pretrained layers.
  $x=1$ represents loading the embedding layer only, $x=3$ represents that plus the first two encoder layers, and $x=23$ represents all layers except for one decoder layer.
  Performance of $x=0$ (i.e. $\RandInit$) and $x=24$ (i.e. $\BGBGInit$) are shown with dashed lines.
  Error bars represent 95\% confidence intervals as reported by the ROUGE script.}
\label{figure:layerwise}
\end{figure}

\subsubsection{Partial pretraining}
In this section, we present the results of an ablation experiment in which we examine the effect of partially pretraining our model.
Results are shown in Fig.~\ref{figure:layerwise}.

The common practice with respect to pretraining summarization models is to start from pretrained word-embeddings.
As the results in Fig.~\ref{figure:layerwise} indicate, this type of ``shallow pretraining'' is harmful in the case of BERT word embeddings.
Notably, the $\RandInit$ condition outperforms all initialization schemes that initialize fewer than 50\% of the model layers.
Consequently, these results indicate that using pretrained word-embeddings from a deep model such as BERT are not suitable for stand-alone use;
instead, deep initialization (i.e., multiple layers of the encoder and decoder are set to pretrained weights) is required.

Finally, we report a strong positive correlation between the \%-initialization of the model layers and the ROUGE-L F1 score.
The Pearson's \textit{r} correlation between them is .8698 (based on the numbers used to create Fig.~\ref{figure:layerwise}).

% Discussion
\section{Discussion}
In this paper, we show that full-network parameter initializations for a Transformer-based model, obtained as a result of multi-stage pretraining that includes BERT initialization as a first step, allows us to train abstractive summarization models that improve upon randomly-initialized baselines.

These results are achieved using a simple maximum-likelihood loss (MLE) setup, do not heavily rely on inference-time heuristics, and complement recent modeling advances such as copy-attention.
In addition to having better performance, our two-stage--pretrained models reach their peak score (against a development set) in far fewer epochs, compared to zero- or one-stage pretrained models.
Given that MLE-based training is already superior in both speed and stablity to RL-based methods, these results encourage us to use this efficient and convenient recipe for training high-quality abstractive summarization systems.

%\section*{Acknowledgments}

\bibliography{../bibliography/garcon}
\bibliographystyle{acl_natbib}

%\appendix

%\section{Appendices}
%\label{sec:appendix}

%\section{Supplemental Material}
%\label{sec:supplemental}

\end{document}